\documentclass{article}

\usepackage{arxiv}

\usepackage[utf8]{inputenc} 
\usepackage[T1]{fontenc}    
\usepackage{hyperref}       
\usepackage{url}            
\usepackage{booktabs}       
\usepackage{amsfonts}       
\usepackage{nicefrac}       
\usepackage{microtype}      
\usepackage{lipsum}
\usepackage{graphicx}
\graphicspath{ {./images/} }

\usepackage{graphicx}
\usepackage{textcomp}
\usepackage{xcolor}
\usepackage{hyperref}

\usepackage{amsmath}
\usepackage{wrapfig}
\usepackage{multirow}
\usepackage{tabularx}
\usepackage{multirow}
\usepackage{amsthm}
\usepackage{amssymb}
\usepackage{subfig}
\usepackage{rotating}
\usepackage{comment}

\title{A framework for predicting, interpreting, and improving Learning Outcomes}

\author{
 Chintan Donda \\
   \And
 Sayan Dasgupta \\
  \And
 Soma S Dhavala \\
 \And
 Keyur Faldu \\
 \And
 Aditi Avasthi\\
 \And
 \texttt{\{chintan, sayan, soma.dhavala, k, aditi\}@embibe.com}\\
 Indiavidual Learning Pvt Ltd\\
 Diamond District, HAL Old Airport Rd, Domlur, \\Bengaluru, Karnataka 560008

}

\begin{document}
\maketitle
\begin{abstract}
It has long been recognized that academic success is a result of both cognitive and non-cognitive dimensions acting together. Consequently, any intelligent learning platform designed to improve learning outcomes (LOs) must provide actionable inputs to the learner in these dimensions. However, operationalizing such inputs in a production setting that is scalable is not trivial. We develop an Embibe Score Quotient model (ESQ) to predict test scores based on observed academic, behavioral and test-taking features of a student. ESQ can be used to predict the future scoring potential of a student as well as offer personalized learning nudges, both critical to improving LOs. Multiple machine learning models are evaluated for the prediction task. In order to provide meaningful feedback to the learner, individualized Shapley feature attributions for each feature are computed. Prediction intervals are obtained by applying non-parametric quantile regression, in an attempt to quantify the uncertainty in the predictions. We apply the above modelling strategy on a dataset consisting of more than a hundred million learner interactions on the Embibe platform. We observe that the Median Absolute Error between the observed and predicted scores is 4.58\% across several user segments, and the correlation between predicted and observed responses is 0.93. Game-like "what-if" scenarios are played out to see the changes in LOs, on counterfactual examples. We briefly discuss how a rational agent can then apply an optimal policy to affect the learning outcomes by treating the above model like an Oracle.
\end{abstract}


\section{Introduction}
Outcome-based Learning is gaining prominence in learner-centric educational services\cite{lo_ed}. Rather than focusing on the the approach to learning, outcome-based learning sets the goals on what a learner can accomplish. It is no coincidence that learning outcomes (LOs) are typically defined using action oriented verbs, largely concerned with academic achievements\cite{lo_def}. This pivoting of measuring LOs allows for self-discovery and has the potential to move away from one-size-fits-all learning solutions. However, such a paradigm shift can put undue stress on already over stretched human resources in the education sector. But, thanks to deeper mobile penetration and faster technology adoption rates, online learning platforms can, not only supplement, but also amplify the reach and potential of every stakeholder in the system, especially teachers and students\cite{lo_dhar}. In fact, the opportunity is so huge that we can even revisit the scope of LOs. Studies have shown that besides skill and competency, psychological traits such as a grit and resilience also contribute to success in general, and in academics in particular \cite{flow,psyq_grit}. This leads to the following important question - how do we measure LOs. There are both direct and indirect ways to measure them\cite{lo_price}. Test scores such as GMAT, GRE, JEE, are the most widely used form of direct measurements. Psychometric tests are available to indirectly measure non-cognitive, psychological constructs\cite{psyq_flow,psyq_res}. In this work, we consider Test Scores as a form of an LO measurement except that the LO is now composite in nature. Our thesis is that, an assortment of skills are required to \textit{perform} well on a test and \textit{succeed}. Online learning platforms have the benefit of capturing fine-grained learner interactions. Depending on the depth of instrumentation, they can track various pedagogic and behavioural aspects of the learning process. We set forth to a framework to leverage such fine-grained signals to positively drive LOs, and do this in a way that the student can be an active participant in the decision process.

\section{Related Work}
In \cite{score_Er}, predictive models were used to classify learners based on their engagement, which instructors can act upon \textit{in situ}. Behavioural patterns associated with performance were mined using Matrix Factorization techniques in \cite{score_meh}. Predicting student test scores based on activity was considered in \cite{score_michael} using a linear regression model with features such as \textit{tasks completed} and \textit{number of attempts}, among others. Using gradient-boosted trees, \cite{score_khe} found that factors such as \textit{scheduling} and \textit{content} are important predictors of students success in MOOCs. A critical review of the state of the art papers in predicting student scores (in MOOCs) was undertaken in \cite{score_josh}. One striking observation is that, in the context of MOOCs, there can be multiple definitions of success (or outcomes) measured in terms of \textit{dropout rate},\textit{certification}, and \textit{ final exam grade}. The importance of feature engineering and the need for providing actionable predictions was emphasized. In the next Section, we discuss the details of our approach.

\section{Modeling Framework}
We consider an extensible, end-to-end modeling framework, where multiple models work in concert, feeding forward their results to downstream models and applications. We lay emphasis on feature engineering pipelines, incorporating uncertainty quantification via prediction intervals, and an optimization framework to produce actionable feedback based on the predictive model.        

\subsection{ Network of Models architecture}
\begin{figure}[htb]
 \centering
  \includegraphics[scale=0.35]{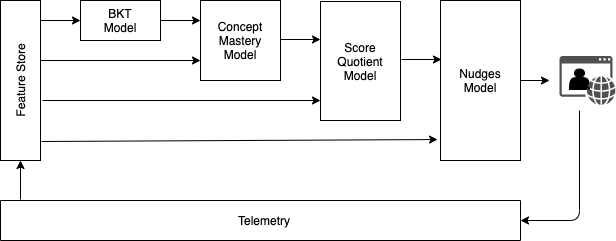}
 \caption{Network of Models}
 \label{fig_arch_01}
 \end{figure}

A functional view of the \emph{Network of Models} architecture is shown in Fig.~\ref{fig_arch_01}. It is inspired by the modality of construction of deep learning networks, where intermediate networks can be seen as feature engineering models. Unlike in traditional deep learning networks, the modules in this architecture can be black box or hand-crafted. 
For example,  we used a Bayesian Knowledge Tracing (\texttt{BKT})\cite{bkt_corbett} model to summarize temporal learner interactions into interpretable features, which are consumed by a Concept Mastery Model. We used Deep Factorization model \texttt{deepFM} \cite{deepfm_guo} to predict academic competency of student across 1,242 concepts. Note that \texttt{Embibe} platform has more than 11,000 concepts in its Knowledge Graph, and for this paper we merged granular concepts into broader concepts (eg: \textit{Differential equation dilution problems} and \textit{Differential Equation Growth and Decay Problems} are mapped to \textit{Application of Differential Equations}) to arrive at the 1,242 concept set. \texttt{deepFM} has the ability to learn complex learner-concept interactions, and has a highly configurable model architecture. Along with the aforementioned academic features, a set of behavioural and test-taking features drive the Embibe Score Quotient model (\texttt{ESQ}) Model. \texttt{ESQ} attempts to model the scoring potential of a learner in a test yet-to-be-taken based on all observations available prior to taking the test\cite{patent_keyur}. The nudges model, using the learnt \texttt{ESQ} as an Oracle, predicts the changes in the learner state (described in terms of the observed features) to achieve  the desired change in the LO. Subsequently we focus on ESQ, and briefly discuss the nudges model.

\subsection{Data Preparation}
\texttt{Embibe} is an online learning platform offering learning solutions for learners through K12 and beyond. It has a collection of immersive content as well as a repository of curated and AI generated assessments and tests. Tens of thousands of practice questions and hundreds of mock tests were made available for learners to prepare and practice \cite{edm_soma}. For the purpose of developing \texttt{ESQ}, we consider data from 175,441 users on the \texttt{Embibe} Platform between January 2017 and December 2019. The raw interactions amount to more than 110 million data points. Data generated from bots and system testers are removed. We also consider only tests where at least 10\% of total questions available should have been attempted, and 10\% of total time available should have been spent.

\subsection{Feature Engineering}
An extensive list of more than 200 features was discussed in \cite{kalyan} useful in MOOC and online platform settings. Pyschological traits such as flow and resilience can also be measured based on survey type questions \cite{psyq_flow,psyq_grit}. On similar lines, we have considered academic, behavioral, effort and test-taking attributes (denoted as AQ, BQ, EQ, TQ respectively), as features to estimate the test score\cite{patent_keyur}. This notion is depicted in Fig~.2.

\begin{figure}[htb]
 \centering
  \includegraphics[scale=0.35]{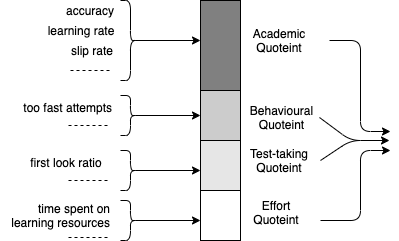}
 \caption{Hypothesized feature categories driving test score}
 \label{fig_feature_02}
 \end{figure}

\textbf{AQ features:}  We use individualized \texttt{BKT}\cite{bkt_gordon}, to summarize the time series attempt level data. Whether a student is in a learned state in a concept/skill, whether the student has a tendency to guess a question against a concept are some of the features given by the \texttt{BKT} model. However, \texttt{BKT} is skill specific and it can not exploit student or concept similarities. In order to leverage this information,  we fit \texttt{deepFM}\cite{deepfm_guo} model to produce student competency across the 1,242 concept set. We reduce the dimension to 50 by random projections\cite{kdd_rp}. A subset of other features considered are \textit{mean accuracy in last three tests}, \textit{score on last test}.


\textbf{TQ features:} Fine-grained attempt-level events are meticulously captured while taking a test on \texttt{Embibe} platform. These events include looking at a question, choosing an answer option, changing an answer option, marking a question for review, attempting a question, swapping subjects, and many others. Using the timestamp for each event, we infer how a student has attempted the test paper at the event level. Each question is also tagged with its ideal time, which is defined as the expected time taken by an idealized student. Based on this we can infer what kinds of questions are attempted too quickly or too slowly. Each question is also tagged as whether \textit{first-look} or not. \textit{First-look} implies that an attempt decision on that question was taken by the student when the student viewed it for the first time, and has not altered that decision in any subsequent visit to that question. Features in TQ bucket summarize question answering behaviour while taking a test. All features are normalized to lie between 0 and 1.

\textbf{BQ features:} In addition to what a learner "knows", scoring on a test also depends on "how" the learner attempts any given test. Traits such as carelessness and over-confidence can affect the final test score. BQ features stand as a proxy for these traits. Features such as \textit{ratio of careless mistakes to all attempts}, \textit{ratio of time spent on non attempts to total time spent}, fall under this bucket.

\textbf{EQ features:} These features measure how much effort was spent by the student between two tests. On the \texttt{Embibe} platform, students can search, browse, watch content, ask questions and perform many other actions. We summarize how students engage based on time spent, and how the student is navigating from one environment to another. Quantification of effort based on what transpired between two events is a useful feature to include\cite{amar}.

\subsection{Model Building}
\begin{figure}[htb]
 \centering
  \includegraphics[scale=0.35]{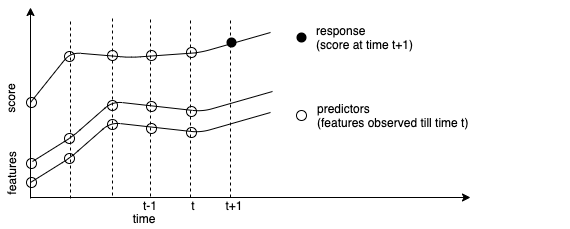}
 \caption{Data flow for Model Setup}
 \label{fig_dataflow_03}
 \end{figure}

Let $y_{t} \in \mathbb{R}$ and $x_{t} \in \mathbb{R}^{p}$  be the response variable (score), and the $p-$dimensional feature vector at time $t$, respectively. Then, we are interested in finding the mapping: $$f: x_{0:t},y_{0:t} \xrightarrow{}  y_{t+1}$$ 
Data flow is depicted in Fig.~\ref{fig_dataflow_03}. 
We consider Random Forests (\texttt{RF})as a baseline model, with 500 trees with depth 5. We only consider $(x_{t}, y_{t})$ to predict $y_{t+1}$ as \texttt{RF} does not naturally accommodate time-series data. We also consider a Recurrent Neural Net (RNN) with stacked LSTM layers. We set the size of the LSTM's hidden state at 20. No hyper parameter tuning was done. 
\subsection{Interpretability}
A model is an approximation of reality. An interpretation can be considered as a way to assert that reality either in quantitative terms or in natural language. It is urged to use interpretable models, where possible\cite{Rudin}. There is a surge in providing interpretations to even seemingly opaque deep learning models. A recently introduced \textit{Shapley values}\cite{shapely} based model explanation technique unifies many existing feature attribution methods such as a \texttt{LIME}, \texttt{DeepLift}, and \texttt{TreeInterpreter}. Shapley possesses some nice theoretical properties such as 1) local accuracy 2) missingness and 3) consistency. In addition, thanks to the additivity property,  feature attributions of any individual instance adds up to the predicted value. In our context, we can break up the test score into AQ, BQ, TQ, EQ components, and provide it as part of engaging feedback. We use \texttt{TreeExplainer} optimized for tree models such as a RFs, and \texttt{DeepExplainer} for RNNs, available in \cite{shapely}.

\subsection{Uncertainty Quantification}
Generally speaking, Machine Learning (ML) algorithms are concerned with predictions. Quantifying uncertainty, either in terms of confidence intervals (CIs) and/or prediction intervals (PIs) is usually an afterthought. The problem is exasperated in the Deep Learning space\cite{uq_nguyen}. Such information is particularly useful when deploying ML models in production settings. For example, the product owner may decide not to use the prediction when it is very vague. Generating CIs for models that use third-party tools are very hard. On the other hand, generating PIs is somewhat straightforward, even though not widely known. It is only recently that Quantile Regression (QR)\cite{Koenker2005} has gained popularity among ML practitioners\cite{takeuchi2006nonparametric,qr_rodrigues}. We use \texttt{sk-garden} to fit Quantile Random Forest, and implement Check Loss, defined below, to fit quantiles with RNNs. The Check Loss  is given as:
    $$ \rho_{\tau}(e) = (\tau - I(e<0))e $$
The non-parametric PIs can be obtained by reporting $(\tau, 1-\tau)^{th}$ quantiles at every observed set of predictors. Not only are PIs easier to implement with third party tools, they are also relatively easier to explain than CIs.

\subsection{From predictions to actions}
\begin{figure}[htb]
 \centering
  \includegraphics[scale=0.5]{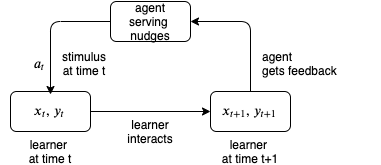}
 \caption{A nudge model}
 \label{fig_nudges_schema}
 \end{figure}

A nudge can be defined as an action taken by an intelligent agent to positively impact LOs. Damgaard et al. \cite{nudges_damgaard} studied the effectiveness of nudges in the education domain, taking examples of nudges tried in different countries to help students study better. Some of their suggested nudges include displaying peer performance, use of relative grading rather than absolute, re-framing incentives as loss, and showing learning videos on the importance of grit and hard work. A schematic of the \texttt{nudges} model is shown in Fig.~\ref{fig_nudges_schema}. 
Providing feedback, a kind of nudge, is also found to be useful \cite{nudges_david}. Bramucci et al \cite{nudges_sherpa} suggested the use of a recommendation engine to help students pick courses that would maximize their success. We can see nudges in the much broader context of Reinforcement Learning (RL). Shayan et al. \cite{ril_shayan} reviewed RL as a generic paradigm for planting nudges into the overall learning framework. In this work, we develop \texttt{ESQ} that can act as an Oracle, which for a given student's state, returns the reward function. This allows us to simulate how students learn, and design optimal policies to provide nudges. That is, we study the inverse problem: \textit{What state should a learner move to, from a given state in order to obtain a particular reward?}. It is akin to back propagating the output gradient (change in reward) to find the input gradient (change in input state). If indeed the causal factors are also encoded in the state, then by solving this inverse problem, we can also suggest which nudge must be given to the student, for the desirable change in the reward (improvement in the test score). We pose this as an optimization problem. Recall that, \texttt{SBT} learns the function $f(.)$, parameterized by $\theta$:  $y \leftarrow f(x;\theta) $. We treat $\theta$ as the unknown while learning the function(training phase), and  while solving for nudges, we treat $\delta x$ as the unknown in the following minimization problem:
$$\min_{\delta x} \ell (y + \delta y, f(x+\delta x; \theta))$$
Here, $\delta y$ is the desired improvement in the score from $y$, $\delta x$ is the anticipated change needed in the student's state from $x$, $\ell(.)$ is a suitable loss function that handles the domain constraints. It is out of scope of this paper to discuss the details, but we demonstrate its utility in the next section.

\section{Results and Discussion}
\begin{figure}[htb]
 \centering
  \subfloat[RF]{{\includegraphics[scale=0.2]{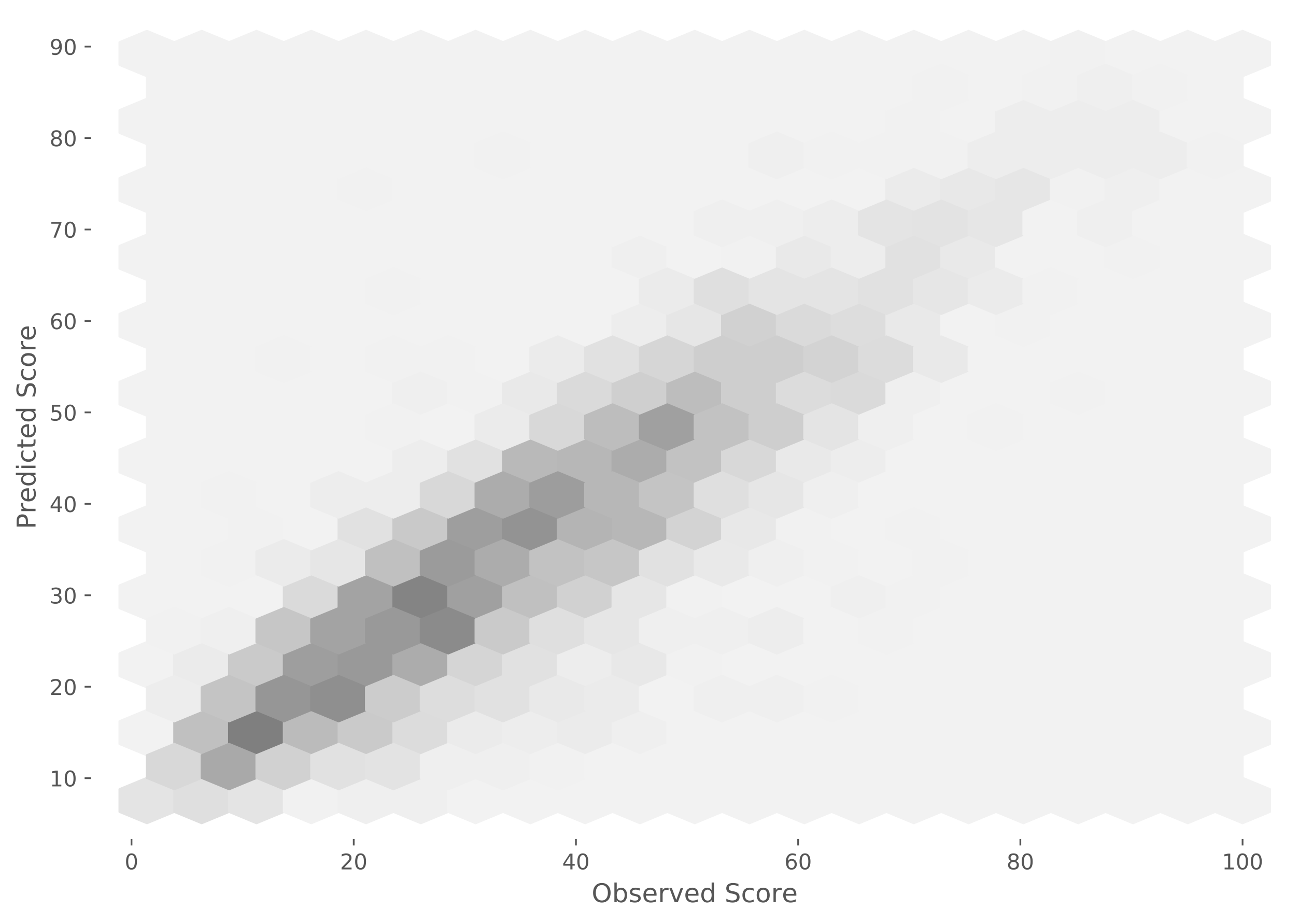} }}%
  \subfloat[RNN]{{\includegraphics[scale=0.2]{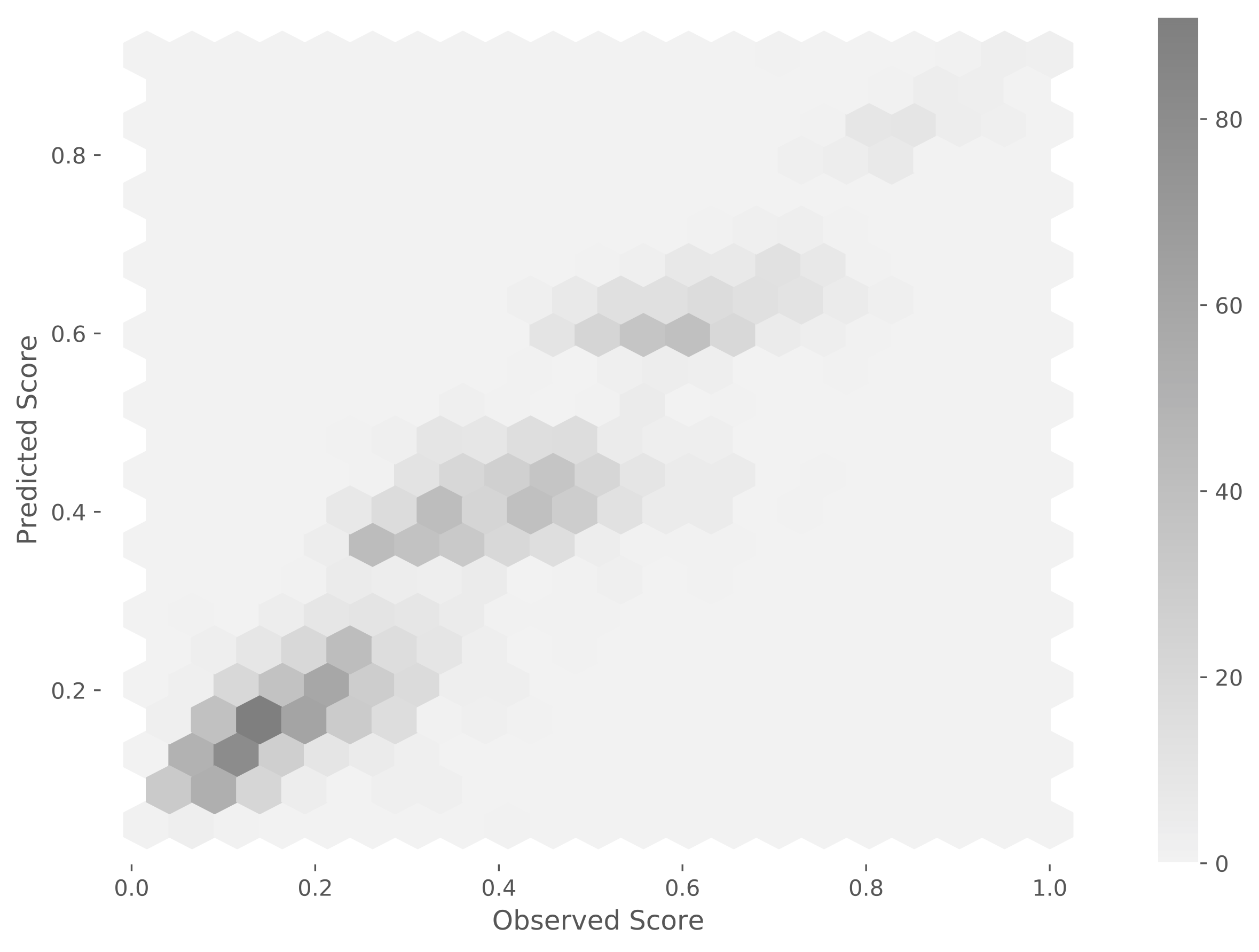} }}%
 \caption{Observed vs Predicted Density}
 \label{fig_residuals_RF_RNN}
 \end{figure}

\begin{figure}[htb]
 \centering
  \subfloat[RF]{{\includegraphics[scale=0.15]{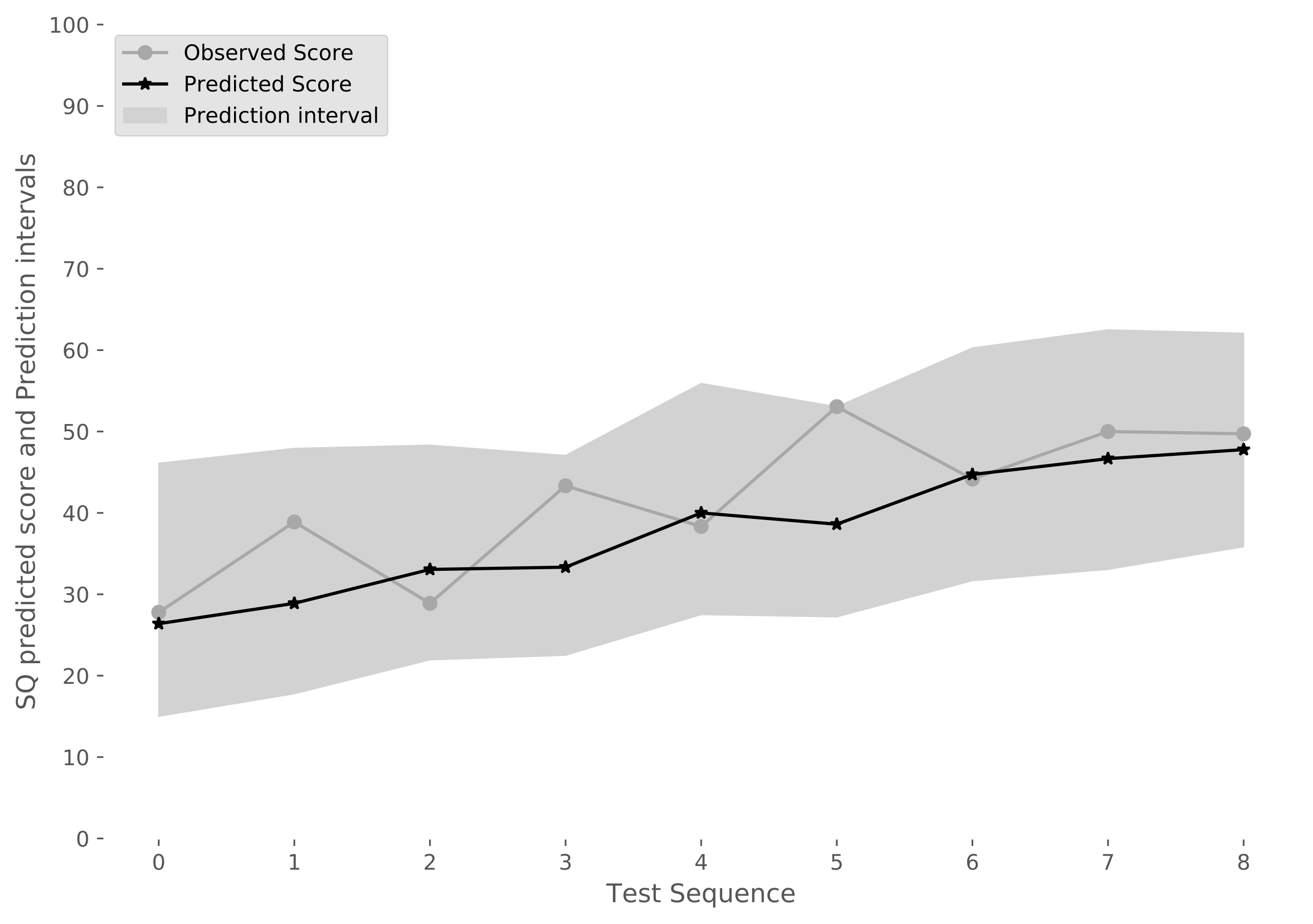} }}%
  \subfloat[RNN]{{\includegraphics[scale=0.15]{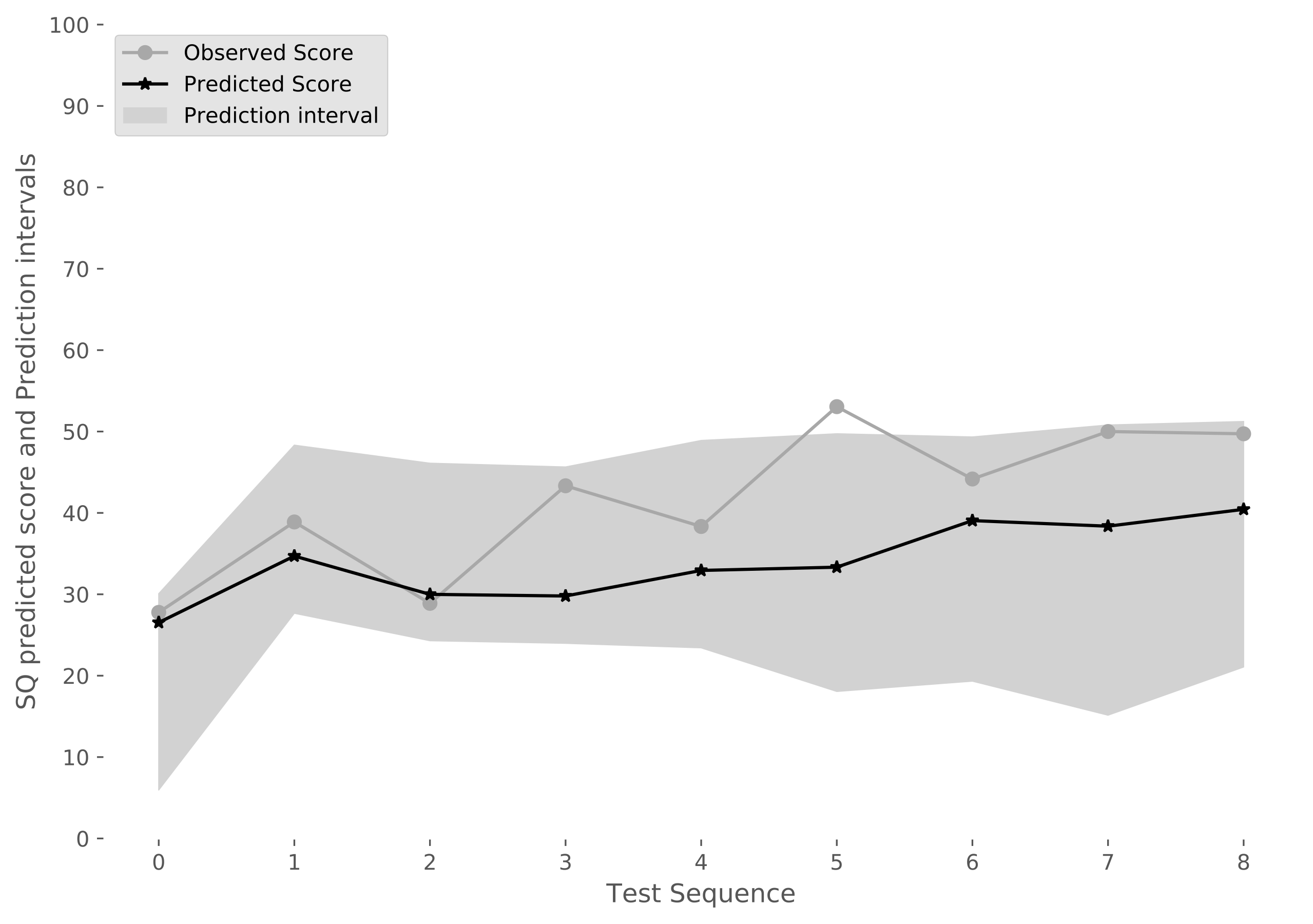} }}%
 \caption{90\% Prediction Intervals for a user}
 \label{fig_PI_RF_RNN}
 \end{figure}

We fit a \texttt{RandomForest} model on the entire dataset. However, we observed that variance was changing with observed score in the residuals, particularly in the low scoring buckets (students score less than 25\%), and the Median Absolute Error (MedAE) was above 10\% which was not acceptable for real world deployment. To address this issue, We segment the users into four buckets based on test scores seen on their previous three test attempts, and fit one \texttt{RandomForest} for each score bucket. We did not use any bucketing strategy to fit RNNs. A two dimensional histogram of observed vs. predicted responses is shown in Fig.~\ref{fig_residuals_RF_RNN}. The residuals of the RNN models have some segmentation, which we believe is due to user encoding we did to reflect the score buckets.  
 Between the two models, the RNN had the smaller Median Absolute Error (MedAE) at 4.53\%, and the Pearson correlation $\rho$ between the predicted and observed density was 0.9331 (see Table ~\ref{tab_model_summary_overall}). With bucketwise RFs, the MedAE is 6.15\% and $\rho$ is 0.8235. To see the effect of adding Concept Mastery vectors, we ran an experiment on one third of the dataset. 
\begin{table*}[ht]
\begin{center}
\caption{Performance summary of RF and RNN models. n: number of samples; p: number of features; r: train/test split ratio. ()* represents Model with Concept Mastery vectors}

\begin{tabular}{|c|c|c|c|c|c|c|c|c|c|c|c|}
    \hline
    \multirow {2} {*} {Model} & \multirow {2} {*} {p} &  \multirow {2} {*} {n} & \multirow {2} {*} {r} & \multicolumn{4}{c|}{Train} &  \multicolumn{4}{c|}{Test}  \\ \cline{5-12} 
     ~ & ~ & ~ & ~ & RMSE & MAE & MedAE &  $\rho$ &  RMSE & MAE & MedAE &  $\rho$\\
    \hline
    RF & 67489	& 54 & 0.8	& 11.21 & 8.13 & 6.16	& 82.35 & 12.50 & 9.12 & 6.73 & 77.15\\ 
    \hline
    RF* & 20358	& 104 & 0.95	& 11.50 &	8.13	& 5.83 & 80.40 & 12.08 & 8.84 & 6.67 &	77.07\\ 
    \hline
    RNN	 & 67489 & 54 & 0.8 &  7.1	& 5.51 &	4.53	& 91.31 &	7.43	&5.69	&4.57	&92.05 \\
    \hline
    RNN*	& 20358  & 104 & 0.95 & 8.84	&6.72	&5.37	&88.60	&10.42	&7.54	&5.91	&88.92 \\
    \hline
\end{tabular}
\label{tab_model_summary_overall}
\end{center}
\end{table*}
We obtain 90\% PIs by fitting a Quantile RF model at $\tau=0.05 \text{ and } 0.95$. We train the same RNN model with Check Loss to fit quantiles. For a particular user, PIs over a period of time are shown in Fig.~\ref{fig_PI_RF_RNN}. 

\begin{figure}[htb]
 \centering
 \begin{turn}{270}
  \includegraphics[scale=0.3]{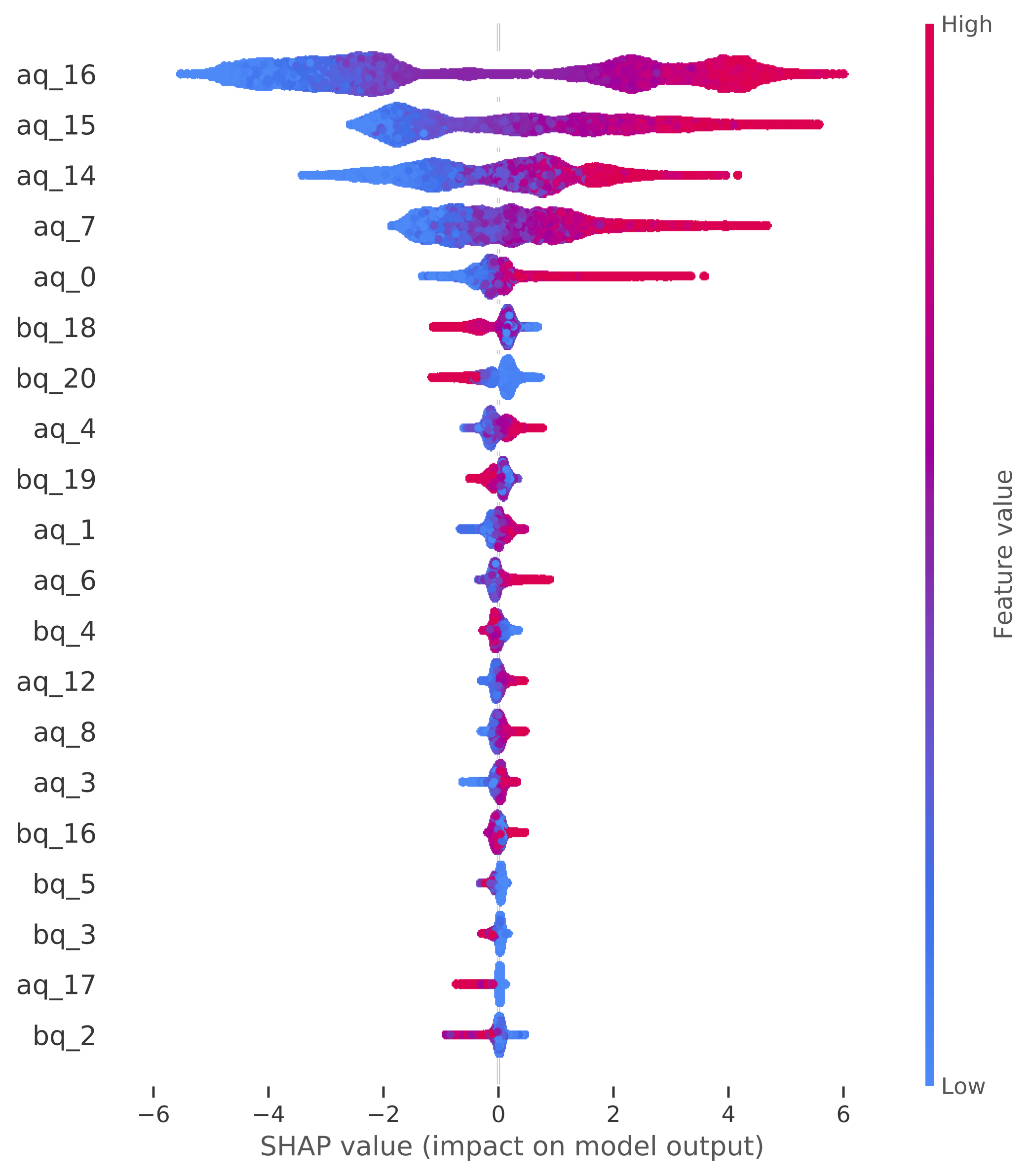}
  \end{turn}
 \caption{Shapley Model Summary}
 \label{fig_Shapley_RF_Model}
 \end{figure}

\begin{figure*}[ht]
 \centering
  \includegraphics[scale=0.4]{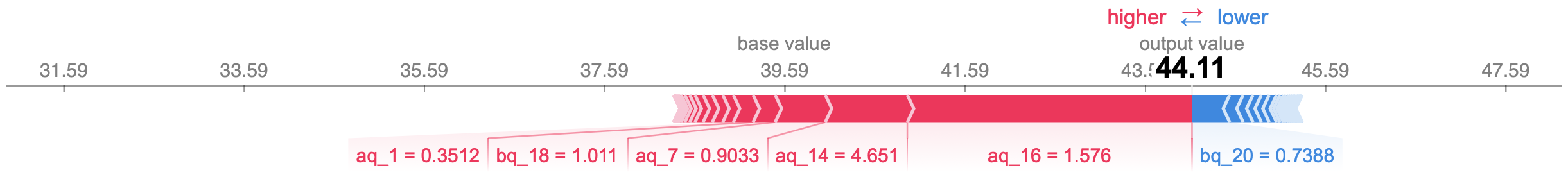}
 \caption{Individualized Shapley values for a user}
 \label{fig_indy_shapley}
\end{figure*}

\begin{figure}[htb]
 \centering
  \includegraphics[scale=0.2]{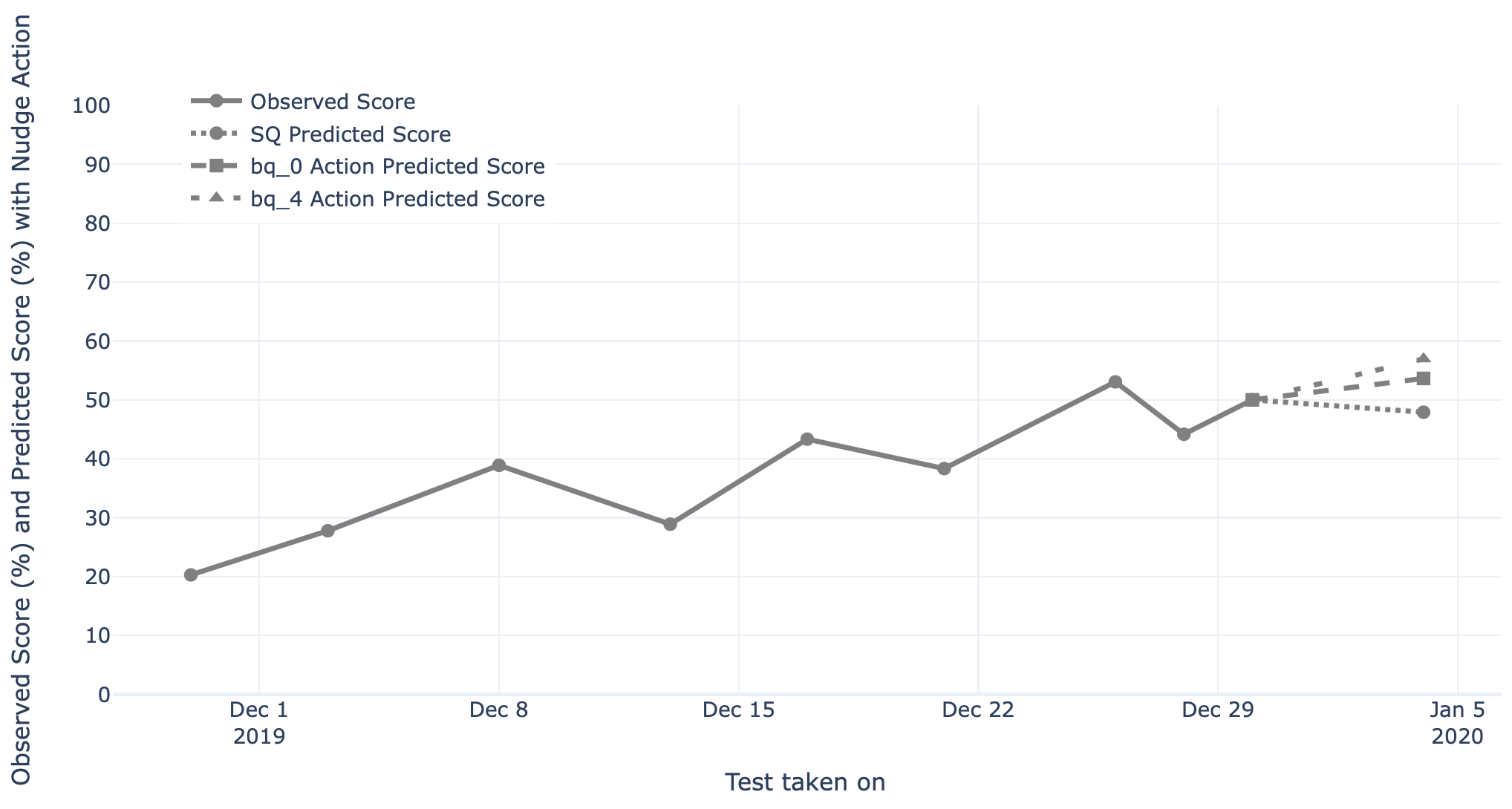}
 \caption{Nudges in Action}
 \label{fig_nudges_eg}
 \end{figure}

 We can use the width of PI as a measure to assert confidence in the predictions -- predictions with wide PIs can be not shown. The Shapley model summary is shown in Fig.~\ref{fig_Shapley_RF_Model}. It turns out that, not surprisingly, \textit{mean accuracy} (coded as \texttt{aq\_16}) is the most discriminating feature. It is interesting to note that, what kind of test a student is taking also influences the test score(\texttt{bq}\textunderscore\texttt{17}). A student with a particular ability can perform differently based on the difficulty of the test - a viewpoint fundamental in Item Response Theory modeling. Overall, AQ features' contribution is 60.98\% to the mean predictions. Feature attributions at an individual level can also be computed based on Shapley values. 
For a particular user, we can show the contribution of each of the features to the score. A force-plot of individualized \texttt{Shapley values} for a user is shown in Fig.~\ref{fig_indy_shapley}. For example, we can see that,  \texttt{aq}\textunderscore\texttt{16} contributed 1.57 points to the predicted score, whereas \texttt{bq}\textunderscore\texttt{20}(\textit{number of test sessions}) reduced predicted score by 0.73 points, possibly because, the student should have practiced this test on par with others. These individualized feature attributions can be used to break up the score into explainable and actionable components.
The nudges model takes this idea further and  systematically perturbs the inputs which can potentially induce a desired a change in the score (score is a proxy for LO in this case). Using the \texttt{nudges} model introduced earlier, we vary one feature at time - much like how coordinate descent works. In fact, the analogy is precisely that. We are optimizing the learning process to improve the LOs. In Fig.~\ref{fig_nudges_eg}, we show affected features which improve the current score by more than 10 points. 

\begin{figure}[htb]
 \centering
  \includegraphics[scale=0.5]{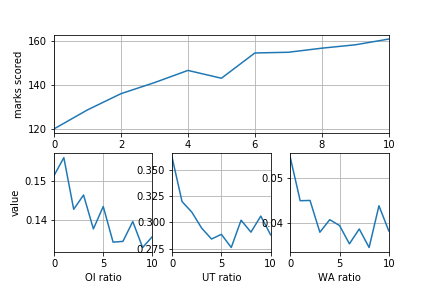}
 \caption{Test-on-test progress of marks scored and change in behavior parameters. Reading clockwise from top are plots for \textit{marks scored, wasted attempt ratio, unused time ratio and overtime incorrects ratio}} 
 \label{fig_test_on_test_change}
 \end{figure}

Our approach draws parallels to recent work on providing explanations to black-box deep learning models via counterfactual examples\cite{dice}. While the \texttt{nudges} model can figure out what the state should be, it does not influence the state directly. On the \texttt{Embibe} platform this happens via a messaging capability - a user can be shown actionable feedback like \textit{You seem to be making careless mistakes. Revise your calculations before submissions}. Once a causal link is established between an action and the affecting feature, \texttt{nudges} model can either influence the learner actions directly or it can be used in a gaming mode, where the learner can play with different simulated actions and pick a suitable action. To see the effect of the feedback, we mined historical data and sampled 1,116 users who have taken at least 10 valid tests (defined earlier) on our platform. In Fig.~\ref{fig_test_on_test_change}, we show how undesirable behavior parameters, such as \textit{wasted attempts, unused time, overtime incorrects}, trend down as marks scored increases test-on-test. The direction of the results is encouraging but we exercise caution in interpreting them. We have not controlled for user attributes, and can not attribute the behavioral change only to the feedback mechanism. However, the analysis suggests that we can persist with the hypothesis that providing useful feedback helps in impacting the intended behavioural aspects. A controlled A/B experiment will test that hypothesis.
\section{Conclusions and Future Work}
In this work, we proposed a framework for predicting Learning Outcomes, defined in terms of test scores. Various features were considered that summarize learner interactions on the platform. We used individualized \texttt{BKT} parameters to summarize temporal attempt level data. Deep Factorization model, with dimensionality reduction, was used to model a user's mastery in 1242 concepts. We are testing our approach on larger datasets. We considered RandomForest model as a baseline to predict test scores. We also considered RNNs as they can naturally handle the sequential nature of the attempt data. While the RNN model gave best results, the residuals did not look smooth. Using \texttt{TreeExplainer}, we obtained individualized Shapley values. Based on this data, we observed that, AQ features contribute about 60.98\% of the scores. In order to fit Shapley values to an RNN, based on \texttt{DeepExplainer}, we had to drop the time-distributed layer, which reduced the RNN performance. We are working on architecture search and feature encodings to improve the performance and model diagnostics, while retaining interpretability. Individualized feature attributions can be presented to provide a breakup of the test score in terms of the features. If the features themselves are actionable, it is possible to directly avail this information to drive actions. We modelled this idea formally by posing nudge recommendation as an optimization problem. Since we are treating the \texttt{ESQ} as an Oracle, many simulations can be performed to identify the optimal policy. Our formalism can be easily incorporated into a Reinforcement Learning paradigm. We are working on the framework to provide optimal nudges and validate this hypothesis.

\section*{Acknowledgements}
The authors would like to thank Achint Thomas for very helpful discussions and Anwar Sheikh for pedagoical suggestions.
\bibliographystyle{abbrv}
\bibliography{sigproc}  

\end{document}